\documentclass[letterpaper]{article} 
\usepackage{aaai2027}  
\usepackage[hyphens]{url}  
\usepackage{graphicx} 
\usepackage{natbib}  
\usepackage{caption} 
\usepackage{algorithm}
\usepackage{algorithmic}
\usepackage{multirow}
\usepackage{pifont}
\usepackage{makecell} 
\usepackage{xcolor}
\definecolor{rowcolor}{RGB}{227,239,255}
\usepackage{tabularx}
\usepackage{amsmath}
\usepackage{tabularx}
\usepackage{booktabs}
\usepackage{colortbl}
\usepackage{xcolor}
\usepackage{soul}
\usepackage{amssymb}
\usepackage{multirow}
\usepackage{makecell}
\definecolor{upcolor}{RGB}{0, 205, 102}
\definecolor{downcolor}{RGB}{250, 128, 114}
\definecolor{rowcolor}{HTML}{FFF2E6}
\definecolor{mygreen}{HTML}{39B64A}
\definecolor{myblack}{HTML}{F0F0F0}
\definecolor{myorange}{HTML}{F77239}

\usepackage{newfloat}
\usepackage{listings}
\DeclareCaptionStyle{ruled}{labelfont=normalfont,labelsep=colon,strut=off} 
\floatstyle{ruled}
\newfloat{listing}{tb}{lst}{}
\floatname{listing}{Listing}

\usepackage{booktabs}

\nocopyright

\title{BUS: Brain-Inspired Unsupervised Self-Reflection \\ via Backward Prediction for Multimodal Reasoning}
\author {
Jiacheng Yang\textsuperscript{\rm 1}
,
 Tongying Xiao\textsuperscript{\rm 1},
  Yunkai Dang\textsuperscript{\rm 1},
   Cong Wang\textsuperscript{\rm 1, \rm 2}, 
   Yuekun Yang\textsuperscript{\rm 1}, 
   Qi Fan\textsuperscript{\rm 1},\\
   Tianyu Ding\textsuperscript{\rm 3},
   Wenbin Li\textsuperscript{\rm 1, \rm4}\corresponding, 
   Feng Miao\textsuperscript{\rm 2},
    Yang Gao\textsuperscript{\rm 1}
}
\affiliations {
    \textsuperscript{\rm 1}State Key Laboratory of Novel Software Technology, Nanjing University, Nanjing, China\\
    \textsuperscript{\rm 2}Institute of Brain-inspired Intelligence, Nanjing University, Nanjing, China\\
    \textsuperscript{\rm 3}Microsoft, Redmond, WA, USA\\
    \textsuperscript{\rm 4}Shenzhen Research Institute of Nanjing University, Shenzhen, China\\
    liwenbin@nju.edu.cn
}

\begin{document}

\maketitle

\begin{abstract}
Current Vision-Language Models (VLMs) often struggle to handle complex visual tasks that require consistent and fine-grained reasoning.
Recent methods aim to train models to facilitate self-reflective reasoning, \textit{i.e.}, reviewing and improving the generated reasoning.
However, they require large volumes of annotated data and lack explicit reflective behavior during test time.
By contrast, humans perform explicit and efficient self-reflection through mechanisms such as backward prediction, \textit{i.e.}, predicting which current states are likely to precede a given future state.
Inspired by neuroscience, this work proposes a novel solution to address these challenges.
We first observe and investigate the phenomenon that mainstream VLMs can perform backward prediction, similar to the human brain.
A label-free training framework named \textit{\textbf{B}rain-inspired \textbf{U}nsupervised \textbf{S}elf-reflection (BUS)} is proposed to leverage and exploit backward prediction capability to enhance reflective reasoning in complex visual tasks.
BUS enables self-verification of reflective reasoning based on backward prediction, providing explicit learning signals under unsupervised conditions.
In this way, BUS eliminates reliance on annotated data while improving reasoning performance.
Designed as a model-agnostic plug-in, our framework is compatible with popular fine-tuning methods, such as Supervised Fine-Tuning (SFT) and Reinforcement Learning (RL). 
Initialized from Qwen3-VL-8B, it improves HR-Bench-8K (+8.0\%),  HR-Bench-4K (+7.7\%), V* Bench (+6.3\%), and MME-RealWorld-Lite (+5.8\%), proving backward prediction is key to advancing reflective reasoning.
\end{abstract}

\section{Introduction}

\begin{figure}[t]
	\centerline{\includegraphics[width=1\columnwidth]{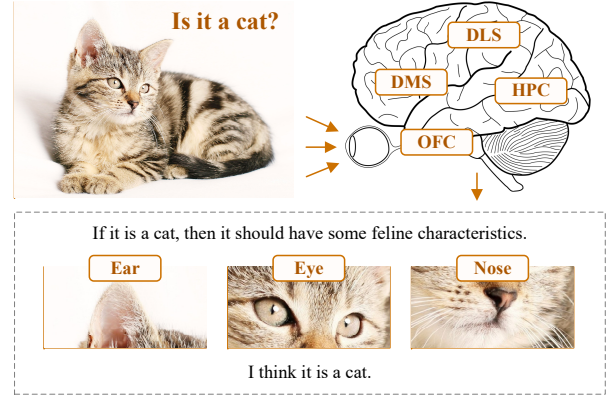}}
	\caption{
	Backward prediction process in question-answering scenarios.
	To answer the given question, the brain predicts which events are likely to precede an image that possibly contains a cat.
	}
	\label{intro}
	\vspace{-6pt}
\end{figure}

Recent breakthroughs in perception and understanding capabilities of Vision-Language Models (VLMs) have shown promise in performing various vision–language tasks, such as visual search~\cite{team2026kimi,openai2025o3,bai2025qwen3} and visual question answering~\cite{fan2026vlm,deepmind2025pro,wang2025internvl3}.
However, improving multimodal reasoning in complex  real-world scenarios still presents significant challenges~\cite{wei2026zooming}.
A primary reason is the presence of inconsistent, unreliable, and incorrect reasoning paths, which negatively affect final performance~\cite{NEURIPS2025_e1d40e92}.
Recent studies propose leveraging self-reflection strategies to boost reasoning performance, enabling VLMs to review and improve their own reasoning process~\cite{yang2026look,shi2026experiential}.
Self-reflection capability promotes logical coherence, deeper understanding, and rigorous reasoning, which are essential for handling complex problems.

Although self-reflection is a highly desirable capability of reasoning VLMs, existing models have been found to perform it inefficiently~\cite{yuan2025agent}.
Cognitive biases in self-reflection strategies do not significantly improve reflective reasoning and can even degrade overall reasoning performance~\cite{zhang-etal-2025-understanding}.
Some recent studies have attempted to enhance the self-reflection capability of reasoning VLMs through fine-tuning, but they still face substantial challenges~\cite{zhou2026v,zhang2026mirror}.
First, they rely heavily on annotated self-reflection datasets, which require large-scale and costly manual annotations~\cite{ding2026learning}.
Second, they struggle to perform explicit reflective behavior during test time, limiting their applicability to complex problems.
After fine-tuning, they typically perform reflective reasoning in a “forward prediction” mode, i.e., generating final responses to questions with little explicit self-reflective behavior.

In contrast, humans seldom rely on a single prediction mode in real-world environments.
Recent neuroscience studies suggest that the brain also performs \textit{backward prediction}, that is, predicting which current states are likely to precede a given future state~\cite{sharp2024humans,DELANGE2026103144}.
They emphasize that backward prediction is a unique and efficient way of making decisions.
Figure~\ref{intro} illustrates an example of the backward prediction process in question-answering scenarios.
Specifically, this process is performed by the coordinated activity of the orbitofrontal cortex (OFC), hippocampus (HPC), dorsolateral striatum (DLS), and dorsomedial striatum (DMS)~\cite{namboodiri2021learning}.
Backward prediction can be understood as a critical form of self-reflection, in which the brain predicts, evaluates, and reviews the reasoning paths preceding its final response.
Borrowing this insight, we propose our research question:
\textit{
Do current VLMs have backward prediction capability?
If so, how can this capability be enhanced to facilitate self-reflective reasoning?
}

To bridge this gap, this work investigates whether VLMs learn backward prediction and how this impacts the decisions they make.
We first design task environments to dissociate different types of prediction and find evidence that current VLMs use backward prediction for decision-making.
Based on this finding, we aim to overcome the two challenges faced by previous self-reflection approaches:
(i) \textit{Reliance on annotated data}: 
backward prediction has the potential to enable models to learn reflection on data without ground-truth labels, thereby achieving label-free training;
(ii) \textit{Lack of explicit reflective behavior}: 
backward prediction can also provide an effective computational mechanism for explicit reflection during test time.

In this work, we propose \textit{\textbf{B}rain-inspired \textbf{U}nsupervised \textbf{S}elf-reflection (BUS)}, a label-free training framework to leverage and exploit backward prediction capability to enhance reflective reasoning.
A key component of this framework is self-verification of reflective reasoning based on both backward and forward prediction.
Given a textual question and an image, {BUS} samples multiple reasoning–answer pairs and guides the model to perform brain-inspired backward prediction, i.e., to predict which of its reasoning paths are likely to precede a known answer.
In this way, the model can perform explicit reflective behavior and be updated on data without access to ground-truth labels.
Different from existing self-reflection approaches, {BUS} directly addresses the need for annotated data and the lack of explicit reflective behavior, offering valuable insights for improving reflective reasoning in complex visual tasks.
Notably, BUS serves as a model-agnostic plug-in that is compatible with popular fine-tuning methods, such as Supervised Fine-Tuning (SFT) and Reinforcement Learning (RL).

In experiments, when applied to post-train Qwen3-VL-8B~\cite{bai2025qwen3}, BUS achieves significant improvements across a range of challenging benchmarks, \textit{i.e.}, $+8.0\%$ on HR-Bench-8K~\cite{wang2025divide}, $+7.7\%$ on HR-Bench-4K~\cite{wang2025divide}, $+6.3\%$ on V* Bench~\cite{Wu_2024_CVPR}, and $+5.8\%$ on MME-RealWorld-Lite~\cite{ICLR2025_df29d63a}.
These improvements are achieved through unsupervised training without any labeled data and further generalize to out-of-distribution tasks.
The main contributions are summarized as follows:

\begin{itemize}
	\item We are the first to investigate the relationship between backward prediction in the human brain and self-reflection in VLMs.
	Experiments verify that VLMs can perform backward prediction for decision-making.
	
	\item We develop BUS, a novel and interpretable framework that utilize backward prediction capability for self-reflection training.
	This approach enhances brain-inspired reasoning without additional manual annotations.
	
	\item We validate our method’s effectiveness on 8 multimodal benchmarks and show that BUS  effectively boosts reasoning performance under unsupervised conditions.

\end{itemize}

\section{Related Work}

The most relevant prior work to our study can be broadly categorized into two research directions:

\textbf{Multimodal Reasoning.} 
Recent advancements in VLMs have marked a watershed moment in the evolution of visual perception and content generation~\cite{team2026kimi,comanici2025gemini,singh2025openai,bai2025qwen3,wang2025internvl3}.
However, real-world scenarios, such as autonomous driving and remote sensing, still remain challenging for most VLMs, as these scenarios often contain complex visual information~\cite{wei2026zooming,wei2026youtu}.
To address this issue, previous works perform image analysis via visual grounding, enabling the model to predict and focus on key image regions before actually answering the question~\cite{ding2026tikart,shi2026improving,wang2025pixel,zhang2025thyme,wang2025traceable}.
They focus only on critical visual information, thereby effectively reducing redundant computations~\cite{zheng2025deepeyes,hong2025deepeyesv2}.
However, despite these advances in image processing, existing approaches often overlook the crucial role of intrinsic reasoning ability.
In this work, we focus on intrinsic reasoning, employing self-reflection to promote logical coherence and deeper visual understanding.

\textbf{Self-reflection. }
Self-reflection allows the model to review and improve its own reasoning process, which becomes essential for improving reasoning quality~\cite{yang2026look,shi2026experiential,pan-etal-2025-lemma,ma-etal-2025-s2r}.
To enable self-reflection, one approach is to directly prompt VLMs to generate and then review their responses~\cite{NEURIPS2025_2c84844a}.
However, studies show that generic prompt instructions often fail to achieve significant intrinsic self-correction and can even lead to performance degradation~\cite{zhang-etal-2025-understanding,yuan2025agent,huang2025beyond}.
Recent works fine-tune VLMs to achieve intrinsic self-correction capability~\cite{zhou2026v,zhang2026mirror,NEURIPS2025_e1d40e92,NEURIPS2025_93397b48,NEURIPS2025_2b0c4fc1}. 
They first construct annotated self-reflection datasets, and then use SFT and RL methods to update model parameters.
These fine-tuning-based methods achieve improved performance compared to direct prompting.
However, they require large-scale datasets and costly manual annotations~\cite{ding2026learning,jian-etal-2025-look}.
Currently, unsupervised training of self-reflection capabilities in VLMs remains largely underexplored.

\begin{figure*}[t]
	\centerline{\includegraphics[width=\textwidth]{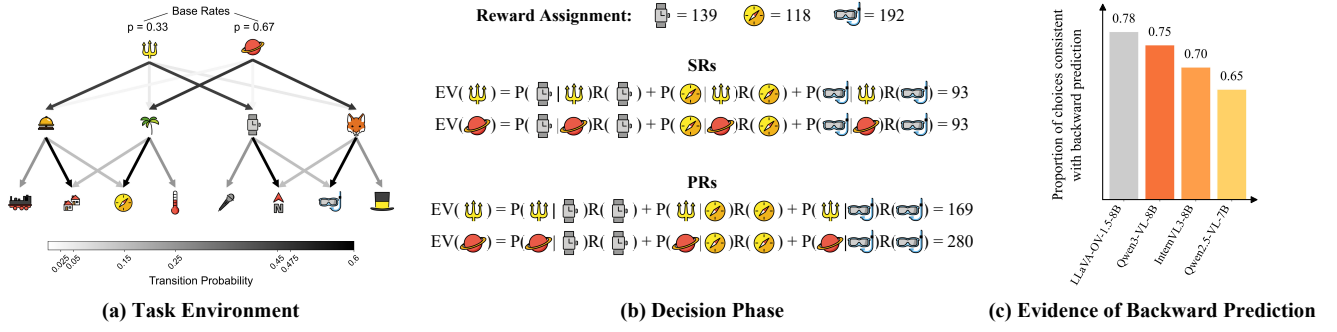}}
	\caption{
	VLMs perform backward prediction.
	(a) Task environment.
	States are represented by images, and darker arrows
	denote higher state-to-state transition probabilities.
	The experiment begins with a learning phase in which all models are presented with state transitions.
	(b) Decision phase.
	Rewards are placed in three states to dissociate backward prediction and forward prediction.
	(c) Evidence of backward prediction.
	Among these models, at least 65\% of the choices are consistent with backward prediction.
	}
	\label{task}
\end{figure*}

In this paper, we focus on training self-reflection capabilities without any external supervision.
Unlike traditional self-reflection methods, {BUS} fine-tunes VLMs on unlabeled data, removing the need for annotated datasets. 
As a novel training paradigm, {BUS} integrates backward prediction mechanism directly into the self-reflection process, which is a more adaptive solution to promote reflective reasoning.

\section{Can VLMs Perform Backward Prediction?}

In this section, we answer the question: do current VLMs have backward prediction capability?
To this end, we conduct pilot experiments designed to verify that VLMs can perform backward prediction, similar to the human brain.
Current neuroscience research suggests that one validation method is to dissociate the two types of predictive behavior based on differences in their representations~\cite{sharp2024humans}.
(i) \textit{Forward prediction}: learning forward prediction can form successor representations (SRs), in which each state is represented as a vector of probabilities specifying the states that typically follow it;
(ii) \textit{Backward prediction}: learning backward prediction can form predecessor representations (PRs), in which each state is represented as a vector of probabilities specifying the states that typically precede it.

\textbf{Experimental Setup.}
We analyze the predictive behavior of popular VLMs, including Qwen3-VL-8B, Qwen2.5-VL-7B, InternVL3-8B, and LLaVA-OneVision-1.5-8B.
We extend the idea of dissociating predictive behaviors from current neuroscience research to VLMs~\cite{sharp2024humans}.
As illustrated in Figure~\ref{task} (a), the experiment is based on a divergent state space containing two starting states, four intermediate states, and eight final states.
The experiment begins with a learning phase in which all models are presented with a starting state and observe transitions from the starting state to an intermediate state, and then from the intermediate state to a final state.
All state transitions have fixed transition probabilities as in Figure~\ref{task} (a).
Notably, we set one starting state to appear more frequently than the other during this phase.

Following the learning phase, we evaluate the VLM’s use of backward prediction using one hundred queries.
In each query, rewards are placed in three intermediate states or final states.
These states are denoted as $\hat{s}_1$, $\hat{s}_2$, and  $\hat{s}_3$, with corresponding rewards $r_1 = R(\hat{s}_1)$, $r_2= R(\hat{s}_2)$, and $r_3= R(\hat{s}_3)$.
Then, all models are asked to choose a starting state to reach the rewards, where the starting states are denoted as $s_1$ and $s_2$.
Under forward prediction, the expected values $V_f$ of the starting states are as follows:

\begin{small}
\begin{equation}
V_f(s_i) = \sum_{k=1}^{3} P(\hat{s}_k | s_i) r_k, \ i \in \left\{ 1, 2 \right\}.
\end{equation}
\end{small}%
Under backward prediction, the expected values $V_b$ of the starting states are as follows:

\begin{small}
\begin{equation}
V_b(s_i) = \sum_{k=1}^{3} P(s_i | \hat{s}_k) r_k = \sum_{k=1}^{3} \frac{P(\hat{s}_k | s_i) P(s_i)}{P(\hat{s}_k)} r_k, \ i \in \left\{ 1, 2 \right\}.
\end{equation}
\end{small}%
By carefully setting the magnitudes and locations of the three rewards in each query, we can distinguish between SR and PR usage if there exist $r_1$, $r_2$, and $r_3$ such that

\begin{small}
\begin{equation}
V_f(s_1) - V_f(s_2) = \sum_{k=1}^{3} \left( P(\hat{s}_k | s_1) - P(\hat{s}_k | s_2) \right) r_k = 0,
\end{equation}
\begin{equation}
V_b(s_1) - V_b(s_2) = \sum_{k=1}^{3} \left( P(s_1 | \hat{s}_k) - P(s_2 | \hat{s}_k) \right) r_k \neq 0.
\end{equation}
\end{small}%
As demonstrated in Figure~\ref{task} (b), under forward prediction the expected values of the two starting states are the same, whereas under backward prediction one of the states has a higher expected value.
Therefore, a preference for the state with higher expected value can be regarded as evidence of backward prediction.

\begin{figure*}[t]
	\centerline{\includegraphics[width=\textwidth]{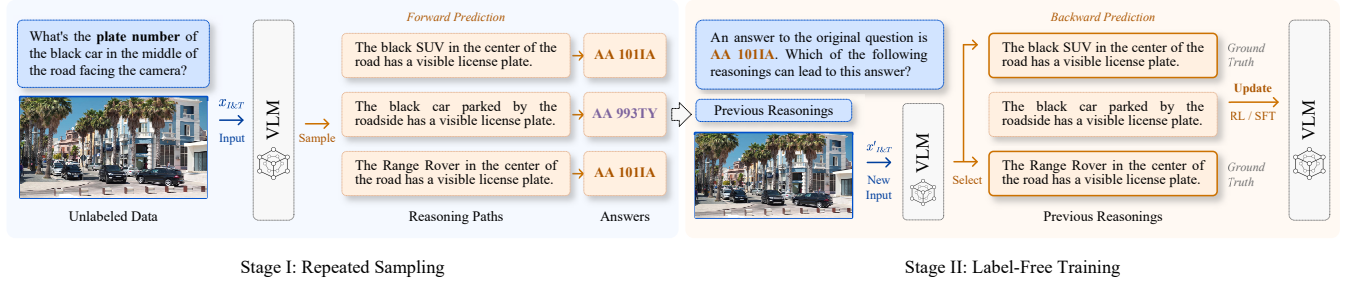}}
	\caption{
	\textbf{BUS Framework.}
	BUS achieves effective label-free training through self-verification of reflective reasoning.
	In Stage I, we generate multiple reasoning–answer pairs through repeated sampling.
	In Stage II, we guide the model to perform brain-inspired backward prediction and fine-tune it on unlabeled data, removing the need for annotated datasets.
	}
	\label{BUS Framework.}
	 \vspace{-6pt}
\end{figure*}

\textbf{Experimental Results.}
We examine the choices made by all VLMs across one hundred queries.
Figure~\ref{task} (c) shows the proportion of choices consistent with backward prediction.
Among these models, at least 65\% of the choices are consistent with backward prediction, providing evidence supporting the hypothesis that the models employ backward prediction for learning and decision-making.
These findings are especially notable because, as a unique and effective form of self-reflection, backward prediction may further enhance the reasoning ability of VLMs.
Our work therefore builds on, and extends, previous neuroscience research showing that humans employ backward prediction using PRs.

\section{Brain-Inspired Unsupervised Self-Reflection}

In this section, we answer the question: how can the backward prediction capability be enhanced to facilitate self-reflective reasoning?
Self-reflection becomes essential for improving reasoning quality.
While previous approaches have attempted to enable self-reflection in VLMs, they rely heavily on annotated datasets and lack explicit reflective behavior~\cite{ding2026learning,NEURIPS2025_e1d40e92}.
To overcome the above limitations, a novel label-free training and plug-in framework named \textit{\textbf{B}rain-inspired \textbf{U}nsupervised \textbf{S}elf-reflection ({BUS})} is proposed to leverage and exploit backward prediction capability to enhance reflective reasoning in complex visual tasks.
As illustrated in Figure~\ref{BUS Framework.}, unlike previous self-reflection approaches, where the model learns in a supervised manner, {BUS} operates on unlabeled data.

\textbf{BUS Framework.}
Given an input $x_{I \& T}$ consisting of an image and a textual question, we first generate multiple reasoning–answer pairs through repeated sampling.
This forward prediction process can be denoted as 
$
\left\{(y_i, a_i)\right\}^n_{i=1}  \sim \pi_\theta (\cdot | x_{I \& T})
$,
where $y_i$ is the $i$-th reasoning, $a_i$ is the $i$-th final answer, and $\pi_\theta$ denotes the model policy parameterized by $\theta$.
We group identical answers into the same category, resulting in a set of categories $\left\{c_j\right\}^m_{j=1}$.
Next, {BUS} guides the model to perform brain-inspired backward prediction, i.e., to predict which of its reasoning paths are likely to precede a known answer.
In particular, we construct a new input $x'_{I \& T}$ based on each answer category $c_j$:
\textit{
Original question: [$x_{I \& T}$].
A model's answer to the original question is: [$c_j$].
Which of the following reasoning(s) can lead to this model's answer?
The choices are listed below: [$y_1 , \dots, y_n$].
}
This backward prediction process can be denoted as 
$
a' \sim \pi_\theta (\cdot | x'_{I \& T})
$,
where $a'$ is the new answer.
Intuitively, the ground truth answer of $x'_{I \& T}$ is the previously sampled reasoning that precedes $c_j$, i.e., $a'_g = \left\{ y_i | a_i = c_j \right\}$.
The model’s backward prediction is considered correct if $a' = a'_g$.
Therefore, by comparing $a'$ and $a'_g$, we can provide explicit learning signals on data without any external supervision.

Compared with traditional self-reflection methods, {BUS} has several significant advantages:
(i) \textit{Label-free training}:
{BUS} can self-verify the accuracy of reflective reasoning based on both backward and forward prediction.
In this way, the model policy is updated on unlabeled data, enabling effective label-free training.
The proposed framework eliminates the cost of manual annotation while promoting the performance of models;
(ii) \textit{On-policy learning}:
BUS can directly fine-tune the model without requiring any additional initialization process.
It learns from dynamic and distribution-shifted inputs, whereas standard self-reflection approaches typically operate in an offline manner. 
Furthermore, the proposed framework is compatible with popular fine-tuning algorithms, such as SFT and RL.

\textbf{SFT-based BUS.}
BUS can be directly used with any SFT algorithm to fine-tune the model $\pi_\theta$ to directly imitate ground-truth $a'_g$ as the response answer.
The loss function is represented as

\begin{small}
\begin{equation}
\mathcal{L}_\text{BUS-SFT}(\theta) = - \sum_{(x'_{I \& T},a'_g)} \text{log} \ \pi_\theta (a'_{g} \vert x'_{I \& T}),
\end{equation}
\end{small}%

\textbf{RL-based BUS.}
We adopt the widely used Group Relative Policy Optimization (GRPO) algorithm as the RL baseline~\cite{guo2025deepseek}.
GRPO first generates a group of $G$ candidate answer $\left\{ a'_i \right\}^G_{i=1}$ and receives corresponding rewards $\left\{ r_i\right\}^G_{i=1}$ through the reward function $R$:

\begin{small}
\begin{equation}
R(a', a'_g) := 
\begin{cases} 
0, & \text{if } a' \not\subseteq a'_g \\
\frac{|a'|}{|a'_g|}, & \text{otherwise}
\end{cases}
\end{equation}
\end{small}%
Thus, the model receives partial credit for selecting a subset of correct reasoning paths and receives zero reward for selecting any incorrect reasoning paths.
The policy is then updated based on the GRPO algorithm.

\begin{table*}[t] 
    
    \small
    \caption{Comparison with competitive methods on popular high-resolution visual benchmarks.
    \textbf{Bold} indicates the best results.
    }
    \newcolumntype{C}{>{\centering\arraybackslash}X}
    \centering
    \begin{tabularx}{\textwidth}{p{2.8cm}|CCCCCCCCCCCC}
    \toprule 
    \multirow{2}{*}{\textbf{Method}}
    & \multicolumn{3}{c}{\textbf{MME-RW-Lite} (ID)}
    & \multicolumn{3}{c}{\textbf{HR-Bench-4K} (OOD)}
    & \multicolumn{3}{c}{\textbf{HR-Bench-8K} (OOD)} 
    & \multicolumn{3}{c}{\textbf{V*} (OOD)}
    \\
    \cmidrule(lr){2-4}
    \cmidrule(lr){5-7}
    \cmidrule(lr){8-10}
    \cmidrule(lr){11-13}
    & \textit{Perc.} & \textit{Reas.} & \textit{Overall} & \textit{FSP} & \textit{FCP} & \textit{Overall} & \textit{FSP} & \textit{FCP} & \textit{Overall} & \textit{Attr.} & \textit{Spa.} & \textit{Overall}\\
    \hline 
    \rowcolor{myblack}\multicolumn{13}{c}{\textit{{Open-source General Models}}} \\
    InternVL3-8B &51.1& 42.9&  47.9&79.3&62.3&70.8& 64.3&59.8&62.0&73.0&71.1&72.3\\
    Qwen2.5-VL-7B & 46.5& 35.9& 42.3&88.8&55.5&72.1& 83.5&  54.0&  68.8&77.4&69.7&74.3\\
    Qwen3-VL-4B  & 51.8& 39.7& 47.1&84.8&62.3&73.5& \textbf{83.5}&50.7&67.1&78.3&69.7&74.9\\
    Qwen3-VL-8B & 54.0 &40.4 & 48.6 & 88.5&56.3&72.4 &81.3 &55.8&68.5&80.2&73.7&77.5\\ 

    \hline 
    \rowcolor{myblack}\multicolumn{13}{c}{\textit{{ Self-reflection Methods}}} \\
    MIRROR 
     &\textemdash&\textemdash&51.5  &\textemdash&\textemdash&72.9  &\textemdash&\textemdash&\textemdash &\textemdash&\textemdash&83.8 \\
    V-Reflection &\textbf{58.5}&45.0&53.9&83.5&61.8&72.6&73.5&58.5&66.3&83.5&78.9&81.7\\
 
    \rowcolor{rowcolor}\textbf{BUS-SFT} (Ours)
    &58.1 &48.1 &54.2 &\textbf{90.0}&58.5&74.3&81.5&56.3&68.9&83.5&\textbf{81.6}&82.7
     \\
    $\Delta$ \textit{vs.} Qwen3-VL-8B
    &\textcolor{myorange}{$\uparrow$ 4.1}
    &\textcolor{myorange}{$\uparrow$ 7.7}
    &\textcolor{myorange}{$\uparrow$ 5.6}
    &\textcolor{myorange}{$\uparrow$ 1.5}
    &\textcolor{myorange}{$\uparrow$ 2.2}
    &\textcolor{myorange}{$\uparrow$ 1.9}
    &\textcolor{myorange}{$\uparrow$ 0.2}
    &\textcolor{myorange}{$\uparrow$ 0.5}
    &\textcolor{myorange}{$\uparrow$ 0.4}
    &\textcolor{myorange}{$\uparrow$ 3.3}
    &\textcolor{myorange}{$\uparrow$ 7.9}
    &\textcolor{myorange}{$\uparrow$ 5.2}
    \\
    \rowcolor{rowcolor}\textbf{BUS-GRPO} (Ours)
        &57.6 &\textbf{49.5} &\textbf{54.4} &\textbf{90.0} & \textbf{70.3}&\textbf{80.1} &82.3 &\textbf{70.8} & \textbf{76.5} &\textbf{85.2} & \textbf{81.6} & \textbf{83.8}\\
        $\Delta$ \textit{vs.} Qwen3-VL-8B &\textcolor{myorange}{$\uparrow$ 3.6}
        &\textcolor{myorange}{$\uparrow$ 9.1}
        &\textcolor{myorange}{$\uparrow$ 5.8}
        &\textcolor{myorange}{$\uparrow$ 1.5}
        &\textcolor{myorange}{$\uparrow$ 14.0}
        &\textcolor{myorange}{$\uparrow$ 7.7}
        &\textcolor{myorange}{$\uparrow$ 1.0}
        &\textcolor{myorange}{$\uparrow$ 15.0}
        &\textcolor{myorange}{$\uparrow$ 8.0}
        &\textcolor{myorange}{$\uparrow$ 5.0}
        &\textcolor{myorange}{$\uparrow$ 7.9}
        &\textcolor{myorange}{$\uparrow$ 6.3}
        \\
    \hline 
    \rowcolor{myblack}\multicolumn{13}{c}{\textit{{Label-based Supervised Methods}}} \\
    PixelReasoner & 53.0&45.6 &49.7 &86.0&60.3& 72.9&80.0& 54.3&66.9&83.5&76.3&80.6\\
    DeepEyes & 55.4&46.8 & 53.2& 91.3& 59.0&75.1&86.8&58.5&72.6&92.1& 86.8&90.0\\
    DeepEyesV2 &54.8 & 48.0&52.4 & 92.8& 63.0&77.9&88.5& 59.0&73.8& 81.7& 80.3&81.8\\
    Thyme &58.2 &48.7 &54.4 &91.0&63.0&77.0&86.5&57.5&72.0&83.5&80.3& 82.2\\
    TreeVGR &58.2 &49.7 &54.9 &89.5& 64.8&77.1&86.0&59.5& 72.8&89.5&84.2&87.4\\
    TikArt &60.5&53.5&57.0&93.8&70.8&82.3&84.5&68.3&76.4&91.3&86.8&89.5\\
    SIEVE&58.4&53.2&56.4&90.0&73.0&81.5&83.0&73.5&78.3&88.7&86.8&88.0\\
    
    \hline 
    \rowcolor{myblack}\multicolumn{13}{c}{\textit{{Private Models}}} \\
    GPT-4o &49.8 &53.9 &45.7 &68.8& 59.0& 63.9 &60.8& 60.0& 60.4 &71.3& 69.7& 70.7\\
    GPT-5 &56.9 &55.5 &56.2 &78.0 &77.0 &77.5 &70.5 &72.8 &71.6 &74.8 &79.0 &76.4\\
    GPT-5-nano &46.1 &49.4 &42.8 &69.0 &61.8 &65.4 &65.8 &61.5 &63.6 &58.3 &72.4 &63.9\\
    Gemini-2.5-pro &52.1 &50.7 &53.5 &88.8 &87.0 &87.9 &84.5 &82.8 &83.6 &84.4 &71.1 &79.1\\
    Gemini-2.5-flash &49.7 &47.3 &52.1 &85.0 &77.0 &81.0 &79.0 &73.5 &76.3 & 84.4  &73.7  &80.1\\
    
    \bottomrule
    \end{tabularx}  
    \label{Comparison with competitive methods on popular high-resolution visual benchmarks.}
    \vspace{-6pt}
\end{table*}

\textbf{Theoretical Analysis.}
We provide theoretical insights to understand
the benefits of BUS in promoting reasoning performance.
In our framework, the model predicts possible preceding reasoning paths conditioned on a given answer category $c_j$ and the question $x_{I \& T}$.
According to Bayes’ theorem, the prediction probability can be expressed as follows:

\begin{small}
\begin{equation}
\begin{aligned}
p_\theta(y\mid c_j,x_{I\&T})
&= \frac{
p_\theta(c_j\mid y,x_{I\&T})p_\theta(y\mid x_{I\&T})
}{
p_\theta(c_j\mid x_{I\&T})
} \\
&\propto p_\theta(c_j\mid y,x_{I\&T})p_\theta(y\mid x_{I\&T}).
\end{aligned}
\end{equation}
\end{small}%
This equation indicates that a reasoning path with a high prediction probability should not only be plausible under $x_{I \& T}$, but also support $c_j$.
Therefore, the model learns the consistency relationship between reasoning paths and answer categories.
Furthermore, the sampled reasoning-answer pairs can be used to construct the joint distribution $\hat p(y,c\mid x_{I\&T})$.
Then the optimization objective of BUS is as follows:

\begin{small}
\begin{equation}
\begin{aligned}
\mathcal{L}_{\text{BUS}}(\theta)
&=-\mathbb{E}_{(y,c)\sim \hat p(\cdot,\cdot\mid x_{I\&T})}\log p_\theta(c\mid y,x_{I\&T})\\
&=H_{\hat p}(c\mid y,x_{I\&T}) \\
& + \mathbb{E}_{y\sim \hat p(\cdot\mid x_{I\&T})}D_{\mathrm{KL}}\left(\hat p(c\mid y,x_{I\&T})\Vert p_\theta(c\mid y,x_{I\&T})\right),
\end{aligned}
\end{equation}
\end{small}%
where the first term is determined by the sampled reasoning-answer pairs, and the second term measures the model’s prediction bias.
The detailed derivation is provided in Appendix A.
Minimizing the loss function effectively promotes the trained models to perform correct backward prediction.
In conclusion, BUS derives a clear learning objective for training models on data without ground-truth labels.

\begin{table*}[t] 
    
    \small
    \caption{Comparison with competitive methods on popular general visual benchmarks.
    \textbf{Bold} indicates the best results.}
    
    \newcolumntype{C}{>{\centering\arraybackslash}X}
    \centering
    \begin{tabularx}{\textwidth}{p{4.7cm}c|CCCCC}
    \toprule
    \textbf{Method} & \textbf{Parameters}& \textbf{MathVerse} & \textbf{MathVista} & \textbf{WeMath}& \textbf{MMStar} &\textbf{Average}\\
    \hline 
    \rowcolor{myblack}\multicolumn{7}{c}{\textit{{Open-source General Models}}}  \\
    InternVL3-8B~\cite{zhu2025internvl3} & 8B&39.8&71.6&50.9&55.8&54.5 \\
    Qwen2.5-VL-7B~\cite{bai2025qwen25vltechnicalreport} &7B &46.3&68.2&57.6&59.3&57.9\\
    Qwen3-VL-8B~\cite{bai2025qwen3} & 8B&52.0&65.7&66.0&64.6&62.1\\
    \hline 
    \rowcolor{myblack}\multicolumn{7}{c}{\textit{{Label-free Self-reflection Method}}}  \\
    \rowcolor{rowcolor}\textbf{BUS-7B} (Ours)
    & 7B&48.8 & 71.0&60.6&63.0&60.9\\
    $\Delta$ \textit{vs.} Qwen2.5-VL-7B && \textcolor{myorange}{$\uparrow$ 2.5} &\textcolor{myorange}{$\uparrow$ 2.8}
    &\textcolor{myorange}{$\uparrow$ 3.0}
    &\textcolor{myorange}{$\uparrow$ 3.7}
    &\textcolor{myorange}{$\uparrow$ 3.0}
    \\
    \rowcolor{rowcolor}\textbf{BUS-8B} (Ours)
    & 8B&\textbf{56.2}&\textbf{72.6}&\textbf{71.3}&\textbf{67.1}&\textbf{66.8}\\
    $\Delta$ \textit{vs.} Qwen3-VL-8B && \textcolor{myorange}{$\uparrow$ 4.2} &\textcolor{myorange}{$\uparrow$ 6.9}
    &\textcolor{myorange}{$\uparrow$ 5.3}
    &\textcolor{myorange}{$\uparrow$ 2.5}
    &\textcolor{myorange}{$\uparrow$ 4.7}
    \\
    \hline 
    \rowcolor{myblack}\multicolumn{7}{c}{\textit{{Label-based Self-reflection Methods}}}  \\
    
    SRPO~\cite{NEURIPS2025_e1d40e92} & 7B&55.8&{75.8}&71.6&\textemdash&\textemdash \\
    
    VL-Rethinker~\cite{NEURIPS2025_2c84844a} & 7B&52.9&74.4 & 69.1 &61.9&64.6 \\
    OpenVLThinker~\cite{deng2025openvlthinker} & 7B&45.7&71.2&66.7&63.4&61.8\\
    VLAA-Thinker~\cite{chen2025sft} & 7B&52.7&69.7&70.2&49.7&60.6\\
    Vision-R1~\cite{huang2025vision} & 7B&52.4&70.6&{73.9}&61.4&64.6\\
    AnE~\cite{wang2026ane} &
    7B& 62.3&81.2& \textemdash & 69.9 &\textemdash
    \\
    Solution-back~\cite{yang2026look} & 7B&51.8&72.3&70.8&\textemdash&\textemdash\\
    Octopus~\cite{ding2026learning} &8B&68.5&82.1&84.0&75.2&77.5 \\
%
    \bottomrule
    \end{tabularx}  
    \label{Comparison with competitive methods on popular general visual benchmarks.}
	\vspace{-6pt}
\end{table*}

\section{Experiments}

\subsection{Fine-grained Visual Reasoning}
\textbf{Benchmarks.}
We evaluate the proposed method on several benchmarks targeting high-resolution visual understanding capabilities of VLMs.
(i) The MME-RealWorld dataset~\cite{ICLR2025_df29d63a} comprises challenging visual question-answering pairs, with an average resolution of $2,076 \times 1,434$.
In {BUS}, we perform unsupervised training on the MME-RealWorld dataset without ground-truth labels;
(ii) HR-Bench~\cite{wang2025divide} serves as an out-of-distribution benchmark that evaluates the model’s performance on 4K and 8K images;
(iii) V* Bench~\cite{Wu_2024_CVPR} serves as an out-of-distribution benchmark with an average image resolution of $2,246 \times 1,583$.

\textbf{Baselines.}
We compare BUS with several state-of-the-art baselines.
(i) Open-source general models include InternVL3-8B~\cite{zhu2025internvl3}, Qwen2.5-VL-7B~\cite{bai2025qwen25vltechnicalreport}, and Qwen3-VL series~\cite{bai2025qwen3};
(ii) Self-reflection methods include MIRROR~\cite{zhang2026mirror} and  V-Reflection~\cite{zhou2026v};
(iii) Supervised training methods include PixelReasoner~\cite{wang2025pixel}, Thyme~\cite{zhang2025thyme}, TreeVGR~\cite{wang2025traceable}, TikArt~\cite{ding2026tikart}, SIEVE~\cite{shi2026improving}, and DeepEyes series~\cite{zheng2025deepeyes,hong2025deepeyesv2};
(iv) Private models include GPT series~\cite{hurst2024gpt,singh2025openai} and Gemini series~\cite{comanici2025gemini}.

\textbf{Implementation Details.}
We employ the Transformer Reinforcement Learning framework~\cite{vonwerra2022trl} to enable distributed training and use Qwen3-VL-8B as the base model.
The hyperparameter $n$ is set to $8$.
Training details are shown in Appendix B.

\textbf{BUS performs well on the in-distribution dataset.}
Table~\ref{Comparison with competitive methods on popular high-resolution visual benchmarks.} presents the accuracy comparison between BUS and baselines on MME-RealWorld-Lite.
BUS-SFT and BUS-GRPO achieve accuracies of 54.2\% and 54.4\%, respectively.
The proposed method delivers significant improvements over the base model Qwen3-VL-8B, surpassing the existing self-reflection model.
Notably, TikArt, SIEVE, and our BUS all use Qwen3-VL-8B as the base model. 
However, TikArt and SIEVE rely on ground-truth labels, whereas our BUS is label-free and still  achieves competitive performance, demonstrating the effectiveness of our training framework.
These results suggest that improving self-reflective reasoning is essential for handling challenging  complex visual tasks.

\textbf{BUS generalizes well to out-of-distribution datasets.}
As shown in Table~\ref{Comparison with competitive methods on popular high-resolution visual benchmarks.}, our BUS also outperforms representative open-source models on out-of-distribution tasks.
Compared to the base model Qwen3-VL-8B, the proposed BUS-GRPO achieves remarkable improvements on challenging high-resolution benchmarks, \textit{i.e.}, +7.7\% on HR-Bench-4K, +8.0\% on HR-Bench-8K, and +6.3\% on V*, providing a flexible solution that better adapts to high-resolution real-world scenarios.
The results highlight the strong adaptability of our method across diverse visual tasks.

\begin{figure}[t]
	\centerline{\includegraphics[width=1\columnwidth]{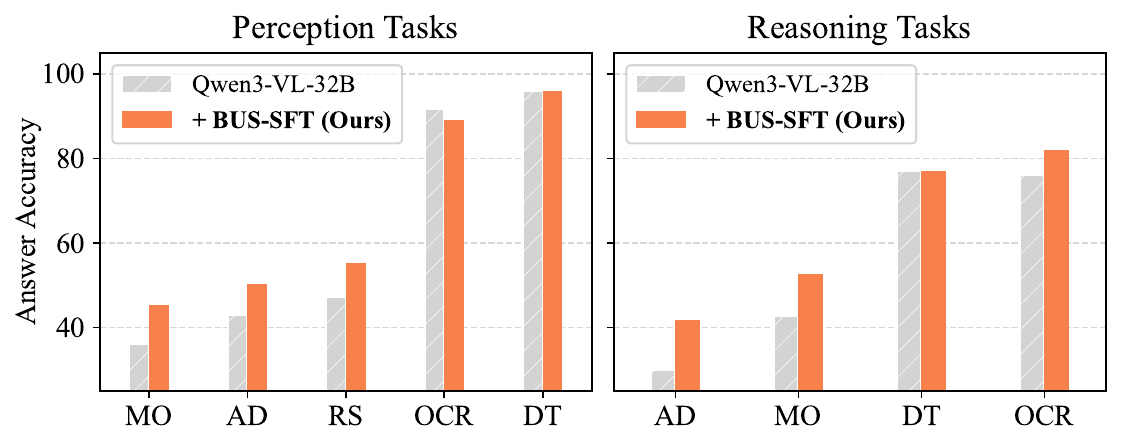}}
	\caption{
	Answer accuracy compared with the larger foundation model Qwen3-VL-32B on MME-RealWorld-Lite.
	Abbreviations: RS-Remote Sensing; MO-Monitoring; DT-Diagram and Table; AD-Autonomous Driving; OCR-Optical Character Recognition in the Wild.
	}
	\vspace{-6pt}
	\label{Qwen3-VL-32B}
\end{figure}

\textbf{BUS naturally scales.}
We use BUS to train the larger foundation model Qwen3-VL-32B.
As shown in Figure~\ref{Qwen3-VL-32B}, we observe significant improvements in most cases as the model size increases (8B $\rightarrow$ 32B), indicating the generalizability of BUS to larger models.
BUS-SFT achieves an accuracy of 58.8\% on MME-RealWorld-Lite, improving upon the base model by 6.8\% through label-free training.

\subsection{Multimodal General Reasoning}

\textbf{Benchmarks.}
In {BUS}, we perform unsupervised training on several general benchmarks without ground-truth labels.
We select MathVerse~\cite{10.1007/978-3-031-73242-3_10}, MathVista~\cite{ICLR2024_663bce02}, and WeMath~\cite{qiao-etal-2025-math} to evaluate mathematical reasoning capabilities. 
MMStar~\cite{NEURIPS2024_2f8ee6a3} is selected to evaluate general reasoning capabilities.

\textbf{Baselines.}
We compare BUS with several state-of-the-art baselines.
(i) Open-source general models include InternVL3-8B~\cite{zhu2025internvl3}, Qwen2.5-VL-7B~\cite{bai2025qwen25vltechnicalreport}, and Qwen3-VL-8B~\cite{bai2025qwen3};
(ii) Self-reflection methods include
SRPO~\cite{NEURIPS2025_e1d40e92}, VL-Rethinker~\cite{NEURIPS2025_2c84844a}, OpenVLThinker~\cite{deng2025openvlthinker}, VLAA-Thinker~\cite{chen2025sft}, Vision-R1~\cite{huang2025vision},
AnE~\cite{wang2026ane}, Solution-back~\cite{yang2026look}, and
Octopus~\cite{ding2026learning}.

\textbf{Performance on Different Foundational Models.}
We apply the proposed method to train Qwen2.5-VL-7B and Qwen3-VL-8B, denoted as BUS-7B and BUS-8B, respectively.
Table~\ref{Comparison with competitive methods on popular general visual benchmarks.} presents the accuracy comparison between BUS and baselines on general benchmarks.
The results show that BUS, trained on data without ground-truth labels, effectively improves overall reasoning performance.
BUS-8B outperforms OpenVLThinker-7B (12k annotations), VLAA-Thinker-7B (55k), Solution-back-7B (15k) across 4 benchmarks.
These results demonstrate the generalizability of BUS across different foundational models.

\begin{figure*}[t]
	\centerline{\includegraphics[width=\textwidth]{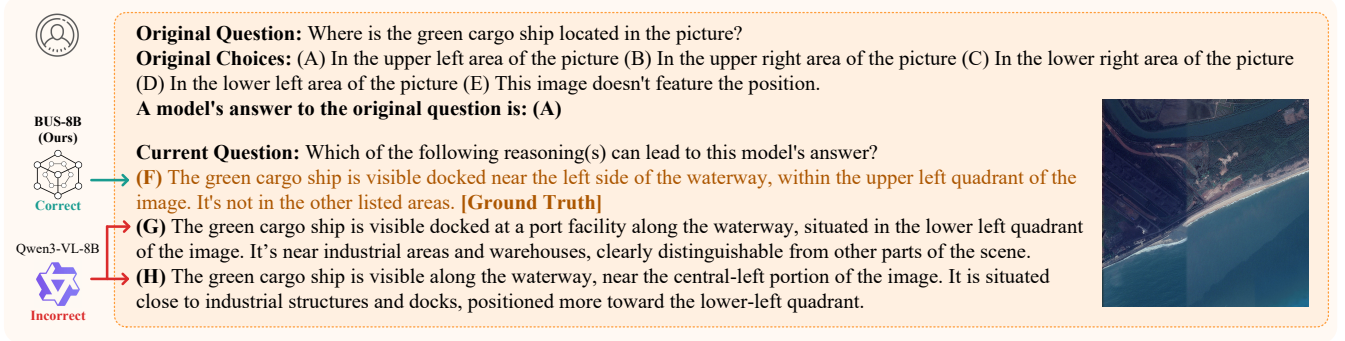}}
	\caption{Visualization results of Qwen3-VL-8B and our BUS-8B on a constructed backward-prediction question.}
	\label{visual}
	\vspace{-6pt}
\end{figure*}

\subsection{Analysis and Discussions}

We present a progressive analysis of the factors enabling BUS to achieve effective visual understanding and reasoning under unsupervised conditions.
The motivation of BUS is to enhance backward prediction capability to facilitate self-reflective reasoning through fine-tuning on data without ground-truth labels.

\begin{figure}[t]
	\centerline{\includegraphics[width=1\columnwidth]{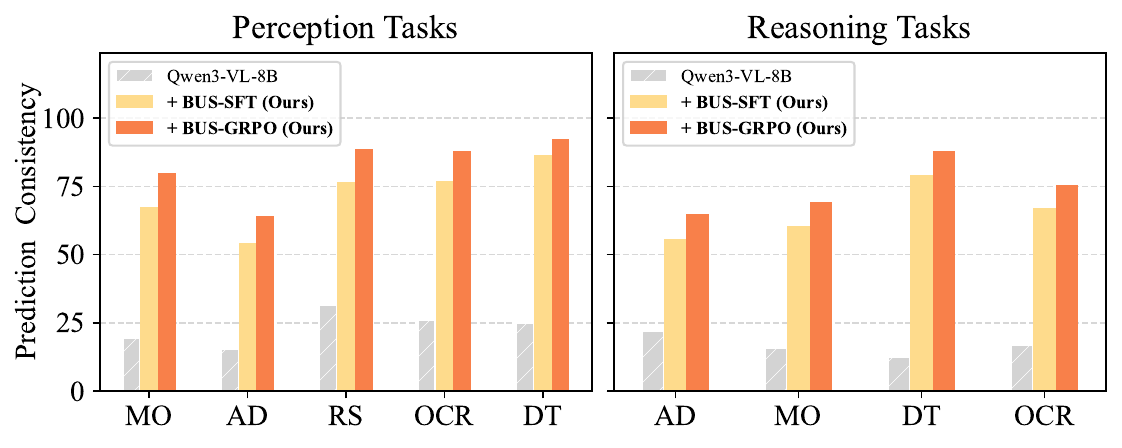}}
	\caption{
	Backward prediction capability compared with the base model Qwen3-VL-8B.
	}
	\vspace{-6pt}
	\label{BackwardPrediction}
\end{figure}

\textbf{Prediction Consistency.}
We first conduct experiments to evaluate the backward prediction capability of different models.
For each question $x_{I \& T}$ in MME-RealWorld, we sample multiple reasoning-answer pairs from Qwen3-VL-8B and then construct a new question $x'_{I \& T}$ for backward prediction.
We compare the prediction consistency of the base model Qwen3-VL-8B, BUS-SFT, and BUS-GRPO on the constructed questions.
Figure~\ref{visual} illustrates a visualization example.
As shown in Figure~\ref{BackwardPrediction}, the proposed method achieves superior backward prediction capabilities.
Compared to the base model, BUS-SFT and BUS-GRPO yield improvements of 38.8\% and 48.6\%, respectively.
The results indicate that backward prediction capability has a positive effect on improving reasoning performance.

\textbf{Training Data.}
A direct difference between BUS and standard self-reflection methods is that BUS involves backward prediction on sampled training data.
The model predicts which of its reasoning paths are likely to precede a sampled answer.
Therefore, a natural question arises: Does BUS remain effective even when the sampled answer is incorrect?
We conduct a set of comparative experiments to investigate the impact of the training data on model performance:
(i) \textit{BUS-GRPO-Incorrect}: 
The training data include only incorrect sampled answers.
(ii) \textit{BUS-GRPO}:
The training data include both correct and incorrect sampled answers.

\begin{table}[t] 
    
    \footnotesize
    \caption{
    Sensitivity analysis on the number of samples $n$.
    }
    \newcolumntype{C}{>{\centering\arraybackslash}X}
    \centering
    \begin{tabularx}{\columnwidth}{p{3.1cm}cCCC}
            \toprule
            \multirow{2}{*}{Method} &\multirow{2}{*}{$n$}& \multicolumn{3}{c}{\textbf{MME-RW-Lite}} \\
            \cmidrule(lr){3-5}
            &  & \textit{Perc.} & \textit{Reas.} & \textit{Overall}\\
            \hline
            Qwen3-VL-8B & - &54.0 &40.4 & 48.6\\
            \hline
            \textit{+ Post-Training} \\ 
            \hline
            \textbf{BUS-GRPO-Incorrect} &8&54.6&44.7&50.7 \\ 
            \multirow{3}{*}{\textbf{BUS-GRPO} }&2&56.5&47.9&53.2\\
            &4&\textbf{57.7}&47.2&53.6\\
            &8&57.6 &\textbf{49.5} &\textbf{54.4}\\
            \bottomrule
        \end{tabularx} 
    \label{ Sensitivity analysis on the number of samples $n$.}
    \vspace{-6pt}
\end{table}

As demonstrated in Table~\ref{ Sensitivity analysis on the number of samples $n$.}, BUS-GRPO-Incorrect still delivers improvements over the base model.
The most fundamental reason lies in the logical association between reasoning paths and answers.
For tasks such as mathematics, even when the final answer is incorrect, it is typically not independent of the mathematical derivations, proofs, and computations in the reasoning process.
Through backward prediction, BUS forces the model to distinguish between logically consistent reasoning and irrelevant or contradictory reasoning.
Consequently, BUS can improve reflective reasoning capability even though the training data include only incorrect sampled answers.
Moreover, the comparison between BUS-GRPO-Incorrect and BUS-GRPO suggests that correct sampled answers provide stronger learning signals.

Next, we present the results of the sensitivity analysis on the number of samples $n$.
We report the answer accuracy of BUS over a range of $n$ (2, 4, and 8) in Table~\ref{ Sensitivity analysis on the number of samples $n$.}.
BUS consistently improves performance under different values of $n$, demonstrating its robustness. 
The results highlight the potential of brain-inspired methods to improve reflective reasoning and overall reasoning performance.

\section{Conclusion}
In this paper, we demonstrate that VLMs employ backward prediction using predecessor representations, similar to humans.
Based on this finding, we propose BUS, a label-free training framework designed for challenging visual tasks.
To reduce reliance on human annotations, BUS enables VLMs to perform self-reflection and self-verification based on both backward and forward prediction, without access to explicit supervision.
Empirical results demonstrate enhanced performance across multiple complex vision-centric tasks.
These results highlight that integrating brain-inspired mechanisms into self-reflection is a promising direction for advancing reasoning capability.
In summary, this paper marks an important direction for reflective reasoning based on backward prediction.
Our contributions aim to provide a foundation for further exploration of unsupervised self-reflection methods.

\bibliography{main}

\end{document}